\documentclass[runningheads]{llncs}
\usepackage[T1]{fontenc}
\usepackage{graphicx}
\usepackage{float}
\usepackage{pgfplots}
\pgfplotsset{compat=1.18}
\usepackage{booktabs}
\usepackage{rotating}
\begin{document}
\title{Refining Filter Global Feature Weighting for Fully-Unsupervised Clustering}
%


\titlerunning{Refining Filter Global FW for Unsupervised Clustering}
%
\author{Fabian Galiș\orcidID{0009-0002-2085-1665} \and
Darian Onchiș\orcidID{0000-0003-4846-3752}}
\authorrunning{F. Galiș and D. Onchiș}
%
\institute{Department of Computer Science, West University of Timisoara, Romania
}
\maketitle              
\begin{abstract}
In the context of unsupervised learning, effective clustering plays a vital role in revealing patterns and insights from unlabeled data. However, the success of clustering algorithms often depends on the relevance and contribution of features, which can differ between various datasets. This paper explores feature weighting for clustering and presents new weighting strategies, including methods based on SHAP (SHapley Additive exPlanations), a technique commonly used for providing explainability in various supervised machine learning tasks. By taking advantage of SHAP values in a way other than just to gain explainability, we use them to weight features and ultimately improve the clustering process itself in unsupervised scenarios.

Our empirical evaluations across five benchmark datasets and clustering methods demonstrate that feature weighting based on SHAP can enhance unsupervised clustering quality, achieving up to a 22.69\% improvement over other weighting methods (from 0.586 to 0.719 in terms of the Adjusted Rand Index). Additionally, these situations where the weighted data boosts the results are highlighted and thoroughly explored, offering insight for practical applications.
\end{abstract}

\section{Introduction}
Clustering is a fundamental task in unsupervised learning that aims to group unlabeled data into meaningful subgroups (clusters) based on similarity measures. It has been applied extensively in a broad range of fields such as image segmentation, customer profiling, and bioinformatics, serving as a primary technique to discover hidden structures in data without relying on labels or annotations \cite{jain2010data,xu2005survey}. Over the decades, a variety of clustering algorithms have been developed—among the most widely known are k-means, hierarchical clustering (e.g., Ward’s method), and density-based algorithms such as DBSCAN \cite{ester1996density,ward1963hierarchical}. Despite their popularity and wide applicability, the performance of clustering algorithms is often highly dependent on the chosen feature representation and dataset, which can significantly affect how the algorithm measures similarities or distances among data points \cite{xu2005survey,jain2010data}.\\

Feature weighting (FW) has emerged as a powerful mechanism to address the sensitivity of clustering algorithms to irrelevant or less informative features. Instead of treating each feature equally, global FW methods assign different weights to features based on their relevance to the clustering objective. There are two general categories of FW techniques, based on the weight estimation strategy:

\begin{enumerate}
    \item \textbf{Filter} FW methods determine weights based on the relationship between the features and a specified reference which corresponds, in an unsupervised scenario, to the intrinsic characteristics of the data \cite{nino2021feature}.

    \item \textbf{Wrapper} FW methods utilize feedback from a given ML algorithm to estimate weights in an iterative, black-box manner. Based on the performance achieved in the previous iteration, which is calculated using either supervised or unsupervised evaluation metrics, the method determines whether to adjust the weights to enhance the model's performance for the next iteration, or not \cite{nino2021feature}.
\end{enumerate}

In addition to FW providing a degree of explainability in regards to feature importance, eXplainable AI (XAI) techniques were developed with that sole purpose in mind, such as SHAP (SHapley Additive exPlanations), which break down each prediction into feature-level contributions by leveraging concepts from cooperative game theory \cite{lundberg2017unified,lundberg2020local}. SHAP has predominantly been used for interpreting and explaining model outputs by assigning each feature a SHAP value, reflecting how much that feature contributed to the final prediction relative to some baseline.

In this paper, we propose a different perspective on SHAP by using it as the foundation for a new approach to FW for clustering. Rather than relying on model-specific interpretations, we exploit the idea behind SHAP values, i.e., quantifying each feature’s contribution, to assign data-driven weights that highlight feature relevance in an unsupervised context. Along with using stand-alone SHAP values as weights, we combine them as an ensemble with other known FW methods. These combinations can exceed not only the performance of the FW methods in question but also the overall performance of unsupervised clustering algorithms.

The paper is structured as follows. Section 2 provides an overview of related work on FW techniques for clustering. Section 3 introduces our methodology, discussing how SHAP values can be adapted for unsupervised FW. Section 4 presents experimental evaluations on multiple datasets, comparing our proposed approach against other FW methods, followed by a discussion subsection. Finally, section 5 concludes the paper and outlines limitations and future potential directions.

\section{Related work}

The use of FW in clustering has been studied extensively. Early work primarily focused on modifying existing clustering algorithms to assign and update feature weights during the clustering process. For instance, in Weighted k-means, each feature is given a weight that is adapted iteratively to minimize within-cluster variance, aiming to place higher emphasis on features that are more discriminative \cite{modha2003feature}.

Similarly, Ward’s method \cite{ward1963hierarchical}, originally introduced for hierarchical clustering, has been extended to account for feature-specific weights (sometimes referred to as Ward variants) by adjusting the distance measure used in building the hierarchy, such as the Minkowski distance \cite{de2015feature}.


Other works have tackled FW by separating it from the clustering procedure itself (i.e., using filter approaches). Filter-based methods rely on statistical tests or correlations to rank features based on their intrinsic properties, which can be representative for potential cluster structures \cite{nino2021feature}.
In \cite{gunes2010kmeans}, a method for feature weighting called K-means Clustering-based Feature Weighting was suggested. This method first extracts features from the frequency domain and calculates their mean, minimum, maximum, and standard deviation as statistical measurements. In the next stage, the K-means algorithm groups these features together, and the average values of the features in relation to the centers of these groups are used as weights.


While these filter FW approaches can be computationally efficient and widely applicable, they do not take into account the behavior of a specific clustering algorithm.
In \cite{nino2021feature}, an extensive classification of FW research works is presented, as well as stating that global filter FW approach in unsupervised learning is not a commonly employed method and few works are encountered in the literature.

In recent years, XAI techniques such as SHAP have transformed how researchers interpret model decisions \cite{lundberg2017unified,lundberg2020local}. SHAP decomposes predictions into additive feature contributions, making it possible to explain complex models in a manner consistent with game-theoretic axioms. Although SHAP has primarily been used in supervised learning settings (classification and regression), its underlying principle—quantifying each feature’s marginal contribution—is promising for providing FW in unsupervised learning. A few studies have started to explore combining XAI methods with cluster analysis, mostly for model interpretability or cluster labeling \cite{yang2021survey}. However, leveraging SHAP directly to derive feature weights that improve clustering outcomes remains largely unexplored territory.

Our work bridges this gap by introducing a SHAP-based global filter FW approach specifically tailored for clustering. By integrating the core ideas of SHAP values into the feature selection and weighting process, we aim to produce meaningful weights that not only enhance clustering performance but also provide what SHAP was meant to offer initially, i.e., feature importance.

\section{SHAP values as feature weights}

\begin{figure}[H]
    \centering
    \includegraphics[width=0.8\linewidth]{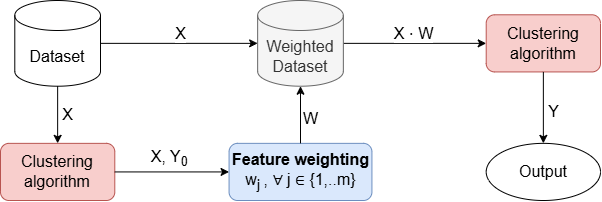}
    \caption{Flowchart of the employed FW methodology}
    \label{fig:flowchart}
\end{figure}

Our primary objective is to integrate the numerical values derived from SHAP into the FW process for clustering tasks. SHAP typically provides a measure of each feature’s marginal contribution to a predictive model’s output in a supervised context. We adapt this concept to unsupervised tasks by training a surrogate predictive model (e.g., a classification model derived from the pseudo-labels) on an initial prediction $Y_0$, made by a clustering algorithm. SHAP values can uncover how much each feature contributes toward distinguishing data points or pseudo-clusters. Once these contributions are computed, they serve as a guideline for assigning the weights for each feature. After transforming the initial data in accordance to the weights $W$, the clustering algorithm can be reapplied, resulting the predictions $Y$, as seen in Fig. \ref{fig:flowchart}. The weighting could be applied again after this step on the labels $Y$ until a certain criterion is met, resulting in a wrapper-like FW method. For the purpose of these experiments, we decided on a single iteration.\\

The resulting SHAP-based weights can also be combined with traditional FW strategies in an ensemble, by multiplying the weights and applying them to data. This ensemble weighting strategy aims to retain each method’s strengths and even surpass the performance of the stand-alone method, as shown in the subsequent section.

\section{Experiments}
In order to run experiments for evaluating the performance of the FW strategies, we employed a hosted T4 GPU provided by Google Colab. For acquiring datasets and implementing algorithms, as well as for using common evaluation metrics, we used the open source library scikit-learn \cite{pedregosa2011scikit}, along datasets from the UCI repository \cite{asuncion2007uci}.

\subsection{Datasets}

We conduct experiments on well-known datasets in the field of machine learning, which can be adapted easily for clustering by ignoring the target feature during the clustering process:

\begin{enumerate}
    \item \textbf{Iris plants dataset} - containing 150 samples of Iris flowers with four features (sepal length, sepal width, petal length, and petal width) and three known classes: Setosa, Versicolour, Virginica \cite{fisher1936use}.
    \item \textbf{Wine recognition dataset} - consists of 178 samples characterized by 13 chemical analysis features of wines derived from three different cultivars, resulting in three classes \cite{aeberhard1992comparison}.
    \item \textbf{Breast cancer Wisconsin (diagnostic) dataset} - featuring 569 samples with 30 features describing tumor cells from clinical samples labeled as benign or malignant, resulting in 2 classes \cite{street1993nuclear}.
    \item \textbf{Optical recognition of handwritten digits dataset} - containing 1797 images of hand-written digits, resulting in 10 classes where each class refers to a digit \cite{kaynak1995methods}.
    \item \textbf{Vehicle Silhouettes} - containing 946 instances for classifying a given vehicle as one of four types, using a set of 18 features extracted from their silhouette \cite{siebert1987vehicle}.
    
\end{enumerate}

Although class labels exist in the datasets, we utilize them only at the evaluation stage to compute external clustering metrics. The clustering itself remains unsupervised.

\subsection{Clustering algorithms}

Four different common clustering algorithms are employed:

\begin{enumerate}
    \item \textbf{\(k\)-means}, a centroid-based algorithm that partitions data into \(k\) clusters by minimizing within-cluster variance \cite{jain2010data};
    \item \textbf{Hierarchical clustering (Ward’s Method)}, a bottom-up approach that successively merges clusters to minimize the increase in sum-of-squares \cite{ward1963hierarchical};
    \item \textbf{Hierarchical Density-Based Spatial Clustering of Applications with Noise (HDBSCAN)}, a density-based clustering algorithm that can handle varying densities, forming a hierarchical tree of possible clusters and extracting stable subclusters \cite{campello2013density};
    \item \textbf{Gaussian Mixture Models (GMM)}, a model-based technique assuming data are generated from a finite mixture of Gaussians, optimized via the Expectation-Maximization (EM) algorithm \cite{xu2005survey}.
\end{enumerate}

\subsection{Feature weighting methods}

In order to use SHAP as a FW method, we first use the initial predictions to train a random forest classifier \cite{breiman2001random} provided by scikit-learn, aggregating the predictions of multiple decision trees. After training, we employ TreeExplainer \cite{lundberg2020local}, an algorithm specifically designed for tree-based models, to calculate SHAP values for the classifier.
The SHAP values, which are computed per sample and per class, are aggregated by taking the mean of their absolute values across both samples and classes. This aggregation yields a single importance score per feature. These scores are then normalized so that they sum to one, providing a set of weights that can be interpreted as the relative importance of each feature.

We integrate and evaluate SHAP as a FW method alongside several other FW or feature selection methods adapted as FW, ensuring that the selected approaches are most varied in terms of their underlying principles, methodologies, and mathematical foundations:

\begin{enumerate}
    \item \textbf{Minkowski distance ($L_p$ norm)}. Inspired by \cite{de2015feature}, this approach can be considered a generalization of both the Euclidean distance ($p=2$) and the Manhattan distance ($p=1$) between two points $x$ and $y$:
    \[
    d(\mathbf{x}, \mathbf{y}) = \left( \sum_{i=1}^{n} \left| x_i - y_i \right|^p \right)^{\frac{1}{p}}
    \]
    
    \item \textbf{Minimum Redundancy Maximum Relevance (mRMR)}. A feature selection method adapted for FW, aiming to maximize the relevance of selected features to the target variable (or pseudo-label in unsupervised cases) while minimizing redundancy among features \cite{peng2005feature}:
    \[
    \max_{S \subseteq F, |S| = k} \left( \frac{1}{|S|} \sum_{f \in S} I(f; c) - \frac{1}{|S|^2} \sum_{f_i, f_j \in S} I(f_i; f_j) \right)
    \]
    where $S$ is the subset of features selected from the complete set $F$, \(I(f; c)\) is mutual information between feature \(f\) and class \(c\), and \(I(f_i; f_j)\) is mutual information between features \(f_i\) and \(f_j\).
    
    \item \textbf{Principal Component Analysis (PCA)}. A technique also used mainly in feature selection and dimensionality reduction. We adapt the principal component loadings as a proxy for feature importance. Larger loadings suggest a stronger influence on the principal components \cite{jolliffe2002principal}.

    \item \textbf{One-way analysis of variance (F-test statistic)}. Often used as a method to compare statistical models, it can be adapted to act as a FW method. It is represented by the ratio of two scaled sums of squares reflecting different sources of variability \cite{guyon2003introduction}.

    \item \textbf{$t$-distributed Stochastic Neighbor Embedding (t-SNE)}. Another unsupervised non-linear dimensionality reduction technique, embedding high-dimensional points in low dimensions in a way that respects similarities between points \cite{vandermaaten08a}.

\end{enumerate}

\subsection{Evaluation Metrics}

We evaluate clustering performance using four metrics that provide a comprehensive and balanced evaluation from different perspectives, both externally and internally:

\begin{enumerate}
    \item \textbf{Adjusted Rand Index (ARI)} measures the similarity between the predicted clusters and ground-truth labels, adjusting for chance \cite{hubert1985comparing};
    \item \textbf{Silhouette Score} quantifies how well samples in the same cluster are similar to each other compared to samples from other clusters \cite{rousseeuw1987silhouettes};
    \item \textbf{Normalized Mutual Information (NMI)} evaluates the amount of mutual information between cluster assignments and ground-truth classes, normalized to the range \([0,1]\) \cite{nino2021feature};
    \item \textbf{Calinski-Harabasz Index (CH)}, also called the Variance Ratio Criterion, assesses the ratio of between-cluster dispersion to within-cluster dispersion \cite{calinski1974dendrite}.
\end{enumerate}

\subsection{Results}

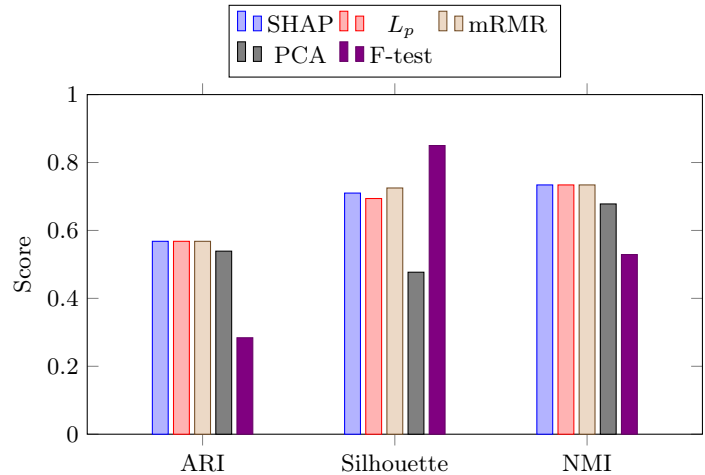
\begin{figure}[ht]
\centering
\begin{tikzpicture}
\begin{axis}[
    ybar,
    width=0.8\textwidth,
    height=0.5\textwidth,
    bar width=6pt,
    enlarge x limits=0.3,
    legend style={
        at={(0.5,1.05)},
        anchor=south,
        legend columns=3,
        font=\small
    },
    ylabel={Score},
    ymin=0,
    ymax=1,  
    symbolic x coords={ARI, Silhouette, NMI},
    xtick=data,
    xticklabel style={}
]

\addplot coordinates {
    (ARI,0.568)
    (Silhouette,0.710)
    (NMI,0.734)
};

\addplot coordinates {
    (ARI,0.568)
    (Silhouette,0.694)
    (NMI,0.734)
};

\addplot coordinates {
    (ARI,0.568)
    (Silhouette,0.725)
    (NMI,0.734)
};

\addplot coordinates {
    (ARI,0.539)
    (Silhouette,0.477)
    (NMI,0.678)
};

\addplot coordinates {
    (ARI,0.284)
    (Silhouette,0.850)
    (NMI,0.529)
};

\legend{SHAP, $L_p$, mRMR, PCA, F-test}

\end{axis}
\end{tikzpicture}
\caption{HDBSCAN metrics across five FW methods on the IRIS dataset.}
\label{fig:iris}
\end{figure}

\begin{figure}[ht]
\centering
\begin{tikzpicture}
\begin{axis}[
    ybar,
    width=0.8\textwidth,
    height=0.5\textwidth,
    bar width=12pt,
    enlarge x limits=0.25,
    legend style={
        at={(0.5,1.05)},
        anchor=south,
        legend columns=3,
        font=\small
    },
    ylabel={Score},
    ymin=0,
    ymax=0.6,  
    symbolic x coords={ARI, Silhouette, NMI},
    xtick=data
]

\addplot coordinates {
    (ARI,0.440)
    (Silhouette,0.233)
    (NMI,0.560)
};

\addplot coordinates {
    (ARI,0.434)
    (Silhouette,0.301)
    (NMI,0.563)
};

\addplot coordinates {
    (ARI,0.423)
    (Silhouette,0.349)
    (NMI,0.532)
};

\legend{SHAP, mRMR, SHAP+mRMR}

\end{axis}
\end{tikzpicture}
\caption{HDBSCAN metrics across three FW methods on the Wine dataset.}
\label{fig:wine}
\end{figure}
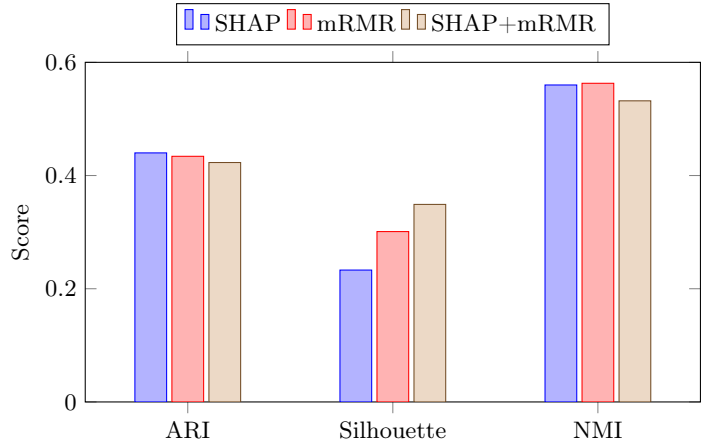

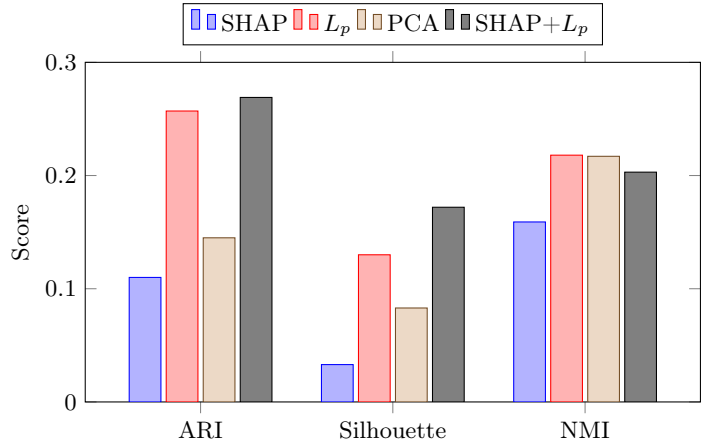
\begin{figure}[ht]
\centering
\begin{tikzpicture}
\begin{axis}[
    ybar,
    width=0.8\textwidth,
    height=0.5\textwidth,
    bar width=12pt,
    enlarge x limits=0.30,
    ylabel={Score},
    ymin=0,
    ymax=0.3,
    legend style={
        at={(0.5,1.05)},
        anchor=south,
        legend columns=4,
        font=\small
    },
    symbolic x coords={ARI, Silhouette, NMI},
    xtick=data,
    xticklabel style={}
]

\addplot coordinates {
    (ARI,0.110)
    (Silhouette,0.033)
    (NMI,0.159)
};

\addplot coordinates {
    (ARI,0.257)
    (Silhouette,0.130)
    (NMI,0.218)
};

\addplot coordinates {
    (ARI,0.145)
    (Silhouette,0.083)
    (NMI,0.217)
};

\addplot coordinates {
    (ARI,0.269)
    (Silhouette,0.172)
    (NMI,0.203)
};

\legend{SHAP, $L_p$, PCA, SHAP+$L_p$}

\end{axis}
\end{tikzpicture}
\caption{HDBSCAN metrics on four FW methods for the Breast cancer dataset.}
\label{fig:cancer}
\end{figure}

To present our results, we plotted the notable situations in Figs \ref{fig:iris}-\ref{fig:cancer} where SHAP-based FW exceeds other FW methods. The CH evaluation metric has been scaled down to fit visually with the other metrics. All results are presented in Tables \ref{tab:iris}-\ref{tab:vehicle} as appendices, with each notable result related to SHAP highlighted. The lines of the result tables where multiple FW methods are enumerated denote a FW ensemble (multiplication).










\subsubsection{Iris plants dataset}

F-test and mRMR frequently emerge as top-performing FW methods across most metrics. SHAP-based weighting is competitive, sometimes outperforming or matching other techniques, particularly in synergy (e.g., SHAP combined with mRMR in HDBSCAN for higher Silhouette). PCA-based weighting is comparatively weaker.

Regarding hierarchical clustering, SHAP yields a respectable ARI of 0.663 and NMI of 0.754, as seen in Table \ref{tab:iris}. The best ARI comes from $L_p$ (0.746) and F-test (0.732). Notably, F-test obtains the highest Silhouette (0.687) and high CH (1022.588).  The ensembles SHAP, $L_p$ and SHAP, mRMR show slightly lower ARIs compared to the best single methods but still remain close in performance. Overall, F-test appears robust for hierarchical clustering, but SHAP remains competitive even when combined with other approaches.

In HDBSCAN, SHAP on its own provides decent performance metrics compared to all other stand-alone FW methods, as seen in Fig. \ref{fig:iris}. When combining SHAP with $L_p$ or mRMR, the Silhouette score increases further (0.747), suggesting that synergy benefits the HDBSCAN distance/neighbor calculations. Interestingly, SHAP combined with F-test leads to a very high CH (9,258.893), but a poor ARI (0.273). This suggests that the Calinski-Harabasz index might not always align with external metrics like ARI in certain data distributions or density-based clustering.

When applying GMM, SHAP is close behind with ARI (0.904) and NMI (0.900). Combining SHAP with mRMR yields an ARI of 0.922, which is still competitive but does not surpass mRMR alone for ARI. SHAP combined with F-test does not exceed F-test alone, indicating no synergy gain in this combination and context.

\subsubsection{Wine recognition dataset}

SHAP-based weighting consistently performs well across \(k\)-means and GMM. Hierarchical clustering sees a stark improvement with the $L_p$ method alone, as visible in Table \ref{tab:wine}.

Regarding \(k\)-means, the ensembles involving SHAP do not strongly surpass the single SHAP approach; in fact, the ensembles have slightly lower ARIs compared to SHAP alone. This might indicate that SHAP’s weighting is already well aligned with the relevant wine features.

When it comes to hierarchical clustering, SHAP alone has an ARI of 0.601, while SHAP combined with PCA jumps to 0.832 in ARI and 0.820 in NMI. This improvement suggests that combining SHAP weights with PCA loadings can better discriminate hierarchical clusters.

In the context of HDBSCAN, mRMR stands out with a high ARI (0.434) and the best Silhouette among single methods (0.301), and SHAP alone performs similarly (0.440 ARI, 0.233 Silhouette), as seen in Fig. \ref{fig:wine}.
The ensemble methods do not yield substantial improvements here in ARI, but do improve Silhouette.

When applying GMM, SHAP has an ARI of 0.947 and NMI of 0.928, which is near the top. The highest ARI across all strategies is from SHAP alone (0.947), while the ensembles of SHAP, $L_p$ and SHAP, PCA remain close but slightly lower in ARI.

\subsubsection{Breast cancer Wisconsin dataset}

As seen in Table \ref{tab:breastcancer}, SHAP and mRMR appear to be the two most reliable strategies across \(k\)-means, hierarchical, and GMM. Both of them combined sometimes helps (particularly with hierarchical clustering) but can harm GMM performance. F-test systematically shows high internal metrics (Silhouette, CH) but fails to align well with external metrics.

When applying \(k\)-means, SHAP performs reasonably well (0.659 ARI, 0.548 NMI). Notably, the SHAP-mRMR ensemble yields a high ARI of 0.671 and a high Silhouette score, of 0.588, suggesting a beneficial combination.

Regarding hierarchical clustering, SHAP performs the best out of the tested methods in terms of ARI (0.719) and NMI (0.599). Moreover, SHAP combined with mRMR stands out with a high ARI of 0.694 and the highest CH (1029.995), indicative of well-separated clusters.

In the context of HDBSCAN, it's notable that the SHAP-$L_p$ ensemble performs the best in terms of ARI (0.269) and CH (72.904), as observed in Fig. \ref{fig:cancer}. Another ensemble, SHAP - F-test, provides the highest Silhouette score (0.353).

Applying GMM, SHAP leads with both ARI (0.793) and NMI (0.682). Just like in the previous algorithm's case, the SHAP F-test ensemble has the highest Silhouette score, 0.634.

\subsubsection{Digits dataset}

As observed in Table \ref{tab:digits}, mRMR tends to dominate $k$-means and HDBSCAN with higher ARI, while SHAP excels for hierarchical clustering.

For other datasets, F-test yields extremely high internal metrics (Silhouette, CH) but very low external agreement (ARI, NMI), suggesting that internal compactness can be misleading when the underlying label distribution is complex.

As this dataset contains 10 clusters, we have opted to omit t-SNE, as the number of components should be fewer than 4 for the Barnes–Hut underlying algorithm of t-SNE to function efficiently.

While combining SHAP weights with other filter/feature selection techniques yielded synergy on some datasets (e.g., IRIS, WINE), it did not consistently improve the digits clustering. Instead, single-method approaches (SHAP or mRMR) frequently performed better, indicating that synergy benefits highly depend on the data distribution.

\subsubsection{Vehicle silhouette dataset}

For this dataset, t-SNE was excluded again, due to the high number of clusters. The results in Table \ref{tab:vehicle} demonstrate that SHAP markedly improves clustering performance across most of the clustering algorithms. Compared to the unweighted dataset, SHAP consistently boosts the ARI and the NMI.

For example, in the case of $k$-means, the unweighted ARI is 0.075 while using SHAP increases it to 0.096. Similar trends are observed for other metrics.

The SHAP and $L_p$ approach, in particular, yields the best overall performance, with an ARI of 0.122 and NMI of 0.168 for $k$-means, an ARI of 0.133 and NMI of 0.214 for hierarchical clustering, respectively 0.168 ARI and 0.334 NMI for the GMM algorithm.

This synergy further emphasizes SHAP’s ability to capture nuanced, potentially non-linear feature importance information which complements the strengths of the $L_p$ norm.

\subsection{Discussion}

In general, SHAP alone delivers often competitive ARI and NMI values across all used datasets, and in some cases surpasses other methods. The synergy with other methods (most notably SHAP-mRMR and SHAP-$L_p$) can improve certain algorithms, but it can also slightly degrade the stand-alone FW methods.

In terms of what algorithms are susceptible to improvement by SHAP alone, the situation is dataset dependent:
\begin{itemize}
    \item Iris dataset - HDBSCAN;
    \item Wine dataset - $k$-means, HDBSCAN, GMM;
    \item Breast cancer dataset - Hierarchical clustering, GMM;
    \item Digits dataset - Hierarchical clustering;
    \item Vehicle silhouette dataset - $k$-means, Hierarchical clustering, GMM.
\end{itemize}

As a practical implication, our findings confirm that no single weighting strategy universally dominates. Rather, performance depends on the synergy between the dataset characteristics, the clustering algorithm, and the metric used. With this in mind, it can be observed that SHAP doesn't underperform, nor does it perform best under one single configuration. Therefore, practitioners can leverage SHAP as a general-purpose, reliable and versatile FW method, while simultaneously gaining insights into each feature's contribution.

\section{Conclusion}

In this paper, we presented feature weighting approaches for clustering, motivated by the need to identify and weight the most informative features during unsupervised learning in a new way. By adapting SHAP—originally designed for supervised settings—we leveraged SHAP values as a principled way to estimate each feature’s contribution in distinguishing pseudo-clusters. Our proposed method was systematically compared against other FW strategies, namely $L_p$, mRMR, PCA, F-test, and t-SNE, both as standalone techniques and in combination with SHAP through ensemble weighting.

Experimental results on standard datasets (Iris, Wine, Breast cancer, Digits and Vehicle Silhouette) and four clustering algorithms ($k$-means, Hierarchical clustering, HDBSCAN, and GMM) demonstrated that SHAP-based feature weighting frequently provides competitive performance, often approaching or outperforming established methods with respect to external clustering metrics like ARI and NMI, especially for data suited for binary clustering, like the Breast cancer dataset. Moreover, in certain scenarios—especially for density-based or hierarchical approaches—combining SHAP with other methods (e.g. SHAP-mRMR or SHAP-$L_p$) proved beneficial in improving cluster separability, as reflected by internal metrics like Silhouette and CH in relatively well-separated clusters, for example the Wine or the Iris dataset. Nonetheless, we observed that these benefits are still dataset- and algorithm-dependent, but perform well enough for this approach to be considered general-purpose and reliable.

Despite promising results, limitations exist. First, deriving SHAP values for clustering involves building a pseudo-supervised setup on unlabeled data (training a model on generated labels), which can increase computational overhead for large datasets. Furthermore, SHAP-based weighting relies on how accurately pseudo-labels approximate underlying cluster structure. If the surrogate model poorly reflects the natural groupings or if the pseudo-labeling process is unstable, the resulting weights may not be optimal. Additionally, while our experiments included multiple well-known datasets, testing on other domains or signal processing data \cite{On1,On2} could further validate robustness and reveal additional edge cases. By addressing these directions, we aim to strengthen the theoretical foundations of SHAP-inspired feature weighting in unsupervised learning and improve its utility in this way, not only as a tool purely for gaining explainability.
\begin{credits}
\subsubsection{\discintname}
The authors have no competing interests to declare that are relevant to the content of this article.
\end{credits}
%
%
%
\bibliographystyle{splncs04}
\bibliography{main}

\appendix

\section*{Appendix 1: Acronym List}
\begin{table}[H]
\centering
\begin{tabular}{ll}
\toprule
\textbf{Acronym} & \textbf{Definition} \\
\midrule
ARI     & Adjusted Rand Index \\
CH      & Calinski-Harabasz (Index) \\
DBSCAN  & Density-Based Spatial Clustering of Applications with Noise \\
FW      & Feature Weighting \\
GMM     & Gaussian Mixture Model \\
HDBSCAN & Hierarchical DBSCAN \\
$L_p$   & $L_p$ norm (Minkowski metric)\\
mRMR    & Minimum Redundancy Maximum Relevance \\
NMI     & Normalized Mutual Information \\
PCA     & Principal Component Analysis \\
SHAP    & SHapley Additive exPlanations \\
Sil     & Silhouette score \\
t-SNE   & $t$-distributed Stochastic Neighbor Embedding\\
XAI     & eXplainable Artificial Intelligence \\
\bottomrule
\end{tabular}
\end{table}

\begin{sidewaystable*}[!ht]
\section*{Appendix 2: Complete list of experiment results}
\centering
\caption{Clustering Results on the IRIS Dataset}
\label{tab:iris}
\renewcommand{\arraystretch}{1.1}  
\setlength{\tabcolsep}{4pt}       
\scriptsize                       
\begin{tabular}{lcccccccccccccccc}
\toprule
 & \multicolumn{4}{c}{\textbf{k-means}} 
 & \multicolumn{4}{c}{\textbf{Hierarchical clustering (Ward)}} 
 & \multicolumn{4}{c}{\textbf{HDBSCAN}} 
 & \multicolumn{4}{c}{\textbf{Gaussian mixture model}} \\
\cmidrule(lr){2-5}\cmidrule(lr){6-9}\cmidrule(lr){10-13}\cmidrule(lr){14-17}
 & \textbf{ARI} & \textbf{Sil} & \textbf{NMI} & \textbf{CH} 
 & \textbf{ARI} & \textbf{Sil} & \textbf{NMI} & \textbf{CH}
 & \textbf{ARI} & \textbf{Sil} & \textbf{NMI} & \textbf{CH}
 & \textbf{ARI} & \textbf{Sil} & \textbf{NMI} & \textbf{CH} \\
\midrule

Unweighted
& 0.433 & 0.480 & 0.590 & 157.360
& 0.615 & 0.447 & 0.676 & 222.719
& 0.564 & 0.495 & 0.718 & 143.161
& 0.517 & 0.475 & 0.657 & 141.232 \\

SHAP
& 0.458 & 0.516 & 0.604 & 224.318
& 0.663 & 0.539 & 0.754 & 449.178
& \textbf{0.568} & 0.710 & \textbf{0.734} & 583.731 
& 0.904 & 0.566 & 0.900 & 667.990 \\

$L_p$
& 0.886 & 0.549 & 0.864 & 592.448 
& 0.746 & 0.590 & 0.798 & 715.563 
& 0.568 & 0.694 & 0.734 & 531.678 
& 0.904 & 0.544 & 0.900 & 596.903 \\

mRMR
& 0.886 & 0.612 & 0.871 & 862.423 
& 0.674 & 0.555 & 0.759 & 720.306 
& 0.568 & 0.725 & 0.734 & 634.911 
& 0.941 & 0.652 & 0.919 & 1097.577\\

PCA
& 0.429 & 0.469 & 0.581 & 156.315 
& 0.590 & 0.454 & 0.622 & 239.008 
& 0.539 & 0.477 & 0.678 & 134.837 
& 0.516 & 0.474 & 0.657 & 140.534 \\

F-test
& 0.886 & 0.727 & 0.864 & 1221.567
& 0.732 & 0.687 & 0.791 & 1022.588
& 0.284 & 0.850 & 0.529 & 306.530  
& 0.886 & 0.727 & 0.864 & 1221.567\\

t-SNE
& 0.676 & 0.473 & 0.700 & 311.005
& 0.720 & 0.487 & 0.784 & 303.726
& 0.568 & 0.657 & 0.734 & 422.920
& 0.904 & 0.419 & 0.900 & 257.122 \\

SHAP, $L_p$
& 0.886 & 0.609 & 0.864 & 929.948 
& 0.720 & 0.617 & 0.784 & 871.219 
& 0.568 & 0.747 & 0.734 & 708.316 
& 0.904 & 0.625 & 0.900 & 1003.069\\

SHAP, mRMR
& 0.886 & 0.649 & 0.864 & 1161.131
& 0.732 & 0.630 & 0.791 & 948.920 
& 0.568 & 0.747 & 0.734 & 702.341 
& 0.922 & 0.654 & 0.898 & 1124.153\\

SHAP, PCA
& 0.688 & 0.489 & 0.706 & 360.396 
& 0.684 & 0.534 & 0.765 & 424.136 
& 0.568 & 0.706 & 0.734 & 566.458 
& 0.904 & 0.559 & 0.900 & 625.727 \\

SHAP, F-test
& 0.886 & 0.727 & 0.864 & 1221.567
& 0.732 & 0.687 & 0.791 & 1022.588
& 0.273 & \textbf{0.889} & 0.530 & \textbf{9258.893}
& 0.886 & 0.727 & 0.864 & 1221.567\\

SHAP, t-SNE
& 0.851 & 0.524 & 0.837 & 514.265
& 0.663 & 0.551 & 0.754 & 497.434
& 0.568 & 0.734 & 0.734 & 650.236
& 0.904 & 0.581 & 0.900 & 741.211 \\

\bottomrule
\end{tabular}

\caption{Clustering Results on the WINE Dataset}
\label{tab:wine}
\renewcommand{\arraystretch}{1.1}
\setlength{\tabcolsep}{4.5pt}
\scriptsize
\begin{tabular}{lcccccccccccccccc}
\toprule
 & \multicolumn{4}{c}{\textbf{k-means}} 
 & \multicolumn{4}{c}{\textbf{Hierarchical clustering (Ward)}} 
 & \multicolumn{4}{c}{\textbf{HDBSCAN}} 
 & \multicolumn{4}{c}{\textbf{Gaussian mixture model}} \\
\cmidrule(lr){2-5}\cmidrule(lr){6-9}\cmidrule(lr){10-13}\cmidrule(lr){14-17}
 & \textbf{ARI} & \textbf{Sil} & \textbf{NMI} & \textbf{CH} 
 & \textbf{ARI} & \textbf{Sil} & \textbf{NMI} & \textbf{CH}
 & \textbf{ARI} & \textbf{Sil} & \textbf{NMI} & \textbf{CH}
 & \textbf{ARI} & \textbf{Sil} & \textbf{NMI} & \textbf{CH} \\
\midrule

Unweighted
& 0.871 & 0.284 & 0.875 & 70.940
& 0.789 & 0.277 & 0.786 & 67.647  
& 0.345 & 0.133 & 0.449 & 24.689  
& 0.897 & 0.284 & 0.875 & 70.940  \\

SHAP
& \textbf{0.880} & 0.435 & 0.850 & 160.129 
& 0.601 & 0.364 & 0.654 & 136.720 
& \textbf{0.440} & 0.233 & 0.560 & 62.690 
& \textbf{0.947} & 0.429 & \textbf{0.928} & 155.921 \\

$L_p$
& 0.790 & 0.370 & 0.770 & 139.341 
& 0.964 & 0.351 & 0.954 & 137.720 
& 0.395 & 0.262 & 0.485 & 57.968 
& 0.852 & 0.365 & 0.836 & 135.390 \\

mRMR
& 0.820 & 0.396 & 0.799 & 153.493 
& 0.817 & 0.394 & 0.794 & 165.233 
& 0.434 & 0.301 & 0.563 & 85.362 
& 0.915 & 0.386 & 0.893 & 146.743 \\

PCA            
& 0.881 & 0.283 & 0.865 & 69.566  
& 0.730 & 0.265 & 0.716 & 64.628  
& 0.341 & 0.130 & 0.425 & 24.450 
& 0.897 & 0.275 & 0.876 & 67.308  \\

F-test          
& 0.374 & 0.591 & 0.453 & 487.088 
& 0.320 & 0.603 & 0.392 & 462.742 
& 0.122 & 0.389 & 0.269 & 25.868 
& 0.318 & 0.599 & 0.394 & 470.792 \\

t-SNE
& 0.698 & 0.322 & 0.706 & 126.626
& 0.837 & 0.312 & 0.815 & 117.914  
& 0.390 & 0.231 & 0.493 & 54.217  
& 0.756 & 0.316 & 0.762 & 122.395  \\

SHAP, $L_p$       
& 0.804 & 0.450 & 0.784 & 195.976 
& 0.588 & 0.415 & 0.652 & 251.179 
& 0.423 & 0.274 & 0.539 & 87.190 
& 0.915 & 0.441 & 0.893 & 186.088 \\

SHAP, mRMR     
& 0.834 & 0.446 & 0.807 & 183.181 
& 0.712 & 0.408 & 0.743 & 208.013 
& 0.423 & 0.349 & 0.532 & \textbf{111.704}
& 0.897 & 0.430 & 0.876 & 171.891 \\

SHAP, PCA      
& 0.880 & 0.433 & 0.850 & 157.875 
& 0.832 & 0.408 & 0.820 & 150.307 
& 0.403 & 0.206 & 0.507 & 50.067 
& 0.931 & 0.428 & 0.909 & 153.549 \\

SHAP, F-test    
& 0.374 & 0.591 & 0.453 & 487.088 
& 0.320 & 0.603 & 0.392 & 462.742 
& 0.104 & \textbf{0.398} & 0.256 & 18.604 
& 0.318 & 0.599 & 0.394 & 470.792 \\

SHAP, t-SNE
& 0.847 & 0.424 & 0.815 & 187.333
& 0.672 & 0.411 & 0.710 & 235.881  
& 0.392 & 0.226 & 0.534 & 65.373  
& 0.895 & 0.422 & 0.882 & 185.763  \\

\bottomrule
\end{tabular}

\caption{Clustering Results on the BREAST CANCER Dataset}
\label{tab:breastcancer}
\renewcommand{\arraystretch}{1.1}
\setlength{\tabcolsep}{4.5pt}
\scriptsize
\begin{tabular}{lcccccccccccccccc}
\toprule
 & \multicolumn{4}{c}{\textbf{k-means}} 
 & \multicolumn{4}{c}{\textbf{Hierarchical clustering (Ward)}} 
 & \multicolumn{4}{c}{\textbf{HDBSCAN}} 
 & \multicolumn{4}{c}{\textbf{Gaussian mixture model}} \\
\cmidrule(lr){2-5}\cmidrule(lr){6-9}\cmidrule(lr){10-13}\cmidrule(lr){14-17}
 & \textbf{ARI} & \textbf{Sil} & \textbf{NMI} & \textbf{CH} 
 & \textbf{ARI} & \textbf{Sil} & \textbf{NMI} & \textbf{CH}
 & \textbf{ARI} & \textbf{Sil} & \textbf{NMI} & \textbf{CH}
 & \textbf{ARI} & \textbf{Sil} & \textbf{NMI} & \textbf{CH} \\
\midrule

Unweighted
& 0.676 & 0.344 & 0.562 & 267.680
& 0.575 & 0.339 & 0.456 & 248.628
& 0.156 & 0.028 & 0.212 & 48.146
& 0.774 & 0.314 & 0.661 & 247.283 \\

SHAP
& 0.659 & 0.575 & 0.548 & 944.123 & \textbf{0.719 }& 0.542 & 0.599 & 880.312 & 0.110 & 0.033 & 0.159 & \textbf{65.027} & \textbf{0.793} & 0.521 & \textbf{0.682} & 752.406 \\
$L_p$           & 0.718 & 0.449 & 0.617 & 495.080 & 0.586 & 0.419 & 0.464 & 411.237 & 0.257 & 0.130 & 0.218 & 56.893 & 0.755 & 0.419 & 0.641 & 447.810 \\
mRMR         & 0.730 & 0.487 & 0.629 & 617.511 & 0.707 & 0.453 & 0.585 & 554.375 & 0.000 & 0.000 & 0.000 & 0.000 & 0.767 & 0.461 & 0.655 & 570.654 \\
PCA          & 0.642 & 0.350 & 0.523 & 269.743 & 0.603 & 0.339 & 0.479 & 258.376 & 0.145 & 0.083 & 0.217 & 33.686 & 0.660 & 0.332 & 0.537 & 247.117 \\
F-test        & 0.126 & 0.618 & 0.070 & 920.864 & 0.127 & 0.611 & 0.071 & 908.811 & 0.001 & 0.352 & 0.046 & 27.081 & 0.137 & 0.633 & 0.082 & 921.494 \\


SHAP, $L_p$     & 0.653 & 0.592 & 0.542 & 1068.392& 0.466 & 0.574 & 0.421 & 900.122 & \textbf{0.269} & 0.172 & 0.203 & \textbf{72.904} & 0.118 & 0.484 & 0.115 & 248.153 \\
SHAP, mRMR   & 0.671 & 0.588 & 0.559 & \textbf{1069.674} & 0.694 & 0.577 & 0.592 & \textbf{1029.995}& 0.000 & 0.000 & 0.000 & 0.000 & 0.174 & 0.456 & 0.159 & 313.117 \\
SHAP, PCA    & 0.659 & 0.573 & 0.548 & 924.684 & 0.689 & 0.546 & 0.568 & 897.500 & 0.076 & 0.054 & 0.145 & 57.197 & 0.103 & 0.513 & 0.114 & 224.856 \\
SHAP, F-test  & 0.126 & 0.618 & 0.070 & 920.864 & 0.127 & 0.611 & 0.071 & 908.811 & 0.001 & \textbf{0.353} & 0.047 & 21.695 & 0.139 & \textbf{0.634} & 0.084 & 918.215 \\

SHAP + t-SNE
& 0.688 & 0.581 & 0.583 & 1004.840
& 0.412 & 0.552 & 0.372 & 766.806
& 0.361 & 0.172 & 0.283 & 88.030
& 0.370 & 0.528 & 0.333 & 590.184 \\

\bottomrule
\end{tabular}
\end{sidewaystable*}

\begin{sidewaystable*}
\caption{Clustering Results on the Digits Dataset}
\label{tab:digits}
\renewcommand{\arraystretch}{1.1}
\setlength{\tabcolsep}{4pt}
\scriptsize
\begin{tabular}{lcccccccccccccccc}
\toprule
 & \multicolumn{4}{c}{\textbf{k-means}} 
 & \multicolumn{4}{c}{\textbf{Hierarchical clustering (Ward)}} 
 & \multicolumn{4}{c}{\textbf{HDBSCAN}} 
 & \multicolumn{4}{c}{\textbf{Gaussian mixture model}} \\
\cmidrule(lr){2-5}\cmidrule(lr){6-9}\cmidrule(lr){10-13}\cmidrule(lr){14-17}
 & \textbf{ARI} & \textbf{Sil} & \textbf{NMI} & \textbf{CH} 
 & \textbf{ARI} & \textbf{Sil} & \textbf{NMI} & \textbf{CH}
 & \textbf{ARI} & \textbf{Sil} & \textbf{NMI} & \textbf{CH}
 & \textbf{ARI} & \textbf{Sil} & \textbf{NMI} & \textbf{CH} \\
\midrule

Unweighted
& 0.530 & 0.135 & 0.672 & 113.060
& 0.664 & 0.125 & 0.795 & 105.825
& 0.209 & 0.041 & 0.580 & 30.762
& 0.546 & 0.117 & 0.690 & 102.577\\

SHAP
& 0.502 & 0.210 & 0.626 & 243.155 
& \textbf{0.700} & 0.181 & 0.786 & 215.512
& 0.376 & 0.005 & 0.650 & 69.313
& 0.518 & 0.170 & 0.651 & 241.091 \\

$L_p$       
& 0.402 & 0.103 & 0.546 & 5.82e+16 
& 0.531 & 0.154 & 0.738 & 184.417
& 0.001 & 0.334 & 0.048 & 30.499
& 0.000 & 1.000 & 0.001 & 3.53e+17 \\

mRMR    
& 0.560 & 0.192 & 0.660 & 201.636 
& 0.667 & 0.185 & 0.788 & 184.384
& 0.454 & 0.083 & 0.724 & 77.951
& 0.615 & 0.199 & 0.717 & 196.626 \\

PCA   
& 0.511 & 0.134 & 0.660 & 115.868 
& 0.660 & 0.136 & 0.778 & 112.902
& 0.219 & 0.032 & 0.584 & 33.854
& 0.557 & 0.128 & 0.687 & 113.244 \\

F-test   
& 0.003 & 0.986 & 0.068 & 49688
& 0.003 & 0.984 & 0.068 & 69182
& 0.003 & 0.992 & 0.070 & 131083
& 0.003 & 0.986 & 0.068 & 50981 \\

SHAP, $L_p$
& 0.441 & 0.199 & 0.566 & 1.38e+12
& 0.635 & 0.185 & 0.730 & 237.468
& 0.001 & 0.976 & 0.042 & 232.357
& 0.000 & 0.000 & 0.000 & 0.000 \\

SHAP, mRMR
& 0.487 & 0.220 & 0.595 & 274.584
& 0.547 & 0.187 & 0.669 & 256.476
& 0.337 & 0.016 & 0.668 & 73.446
& 0.505 & 0.211 & 0.628 & 302.152 \\

SHAP, PCA
& 0.514 & 0.213 & 0.634 & 250.872
& 0.543 & 0.180 & 0.676 & 218.276
& 0.356 & 0.022 & 0.678 & 67.599
& 0.091 & 0.189 & 0.193 & 453.015 \\

SHAP, F-test
& 0.003 & 0.986 & 0.068 & 49688
& 0.003 & 0.984 & 0.068 & 60631
& 0.003 & \textbf{0.993} & 0.070 & \textbf{141957}
& 0.003 & 0.986 & 0.068 & 49688 \\

\bottomrule
\end{tabular}

\caption{Clustering Results on the Vehicle Silhouette Dataset}
\label{tab:vehicle}
\renewcommand{\arraystretch}{1.1}
\setlength{\tabcolsep}{4.5pt}
\scriptsize
\begin{tabular}{lcccccccccccccccc}
\toprule
 & \multicolumn{4}{c}{\textbf{k-means}} 
 & \multicolumn{4}{c}{\textbf{Hierarchical clustering (Ward)}} 
 & \multicolumn{4}{c}{\textbf{HDBSCAN}} 
 & \multicolumn{4}{c}{\textbf{Gaussian mixture model}} \\
\cmidrule(lr){2-5}\cmidrule(lr){6-9}\cmidrule(lr){10-13}\cmidrule(lr){14-17}
 & \textbf{ARI} & \textbf{Sil} & \textbf{NMI} & \textbf{CH} 
 & \textbf{ARI} & \textbf{Sil} & \textbf{NMI} & \textbf{CH}
 & \textbf{ARI} & \textbf{Sil} & \textbf{NMI} & \textbf{CH}
 & \textbf{ARI} & \textbf{Sil} & \textbf{NMI} & \textbf{CH} \\
\midrule

Unweighted
& 0.075 & 0.299 & 0.121 & 390.829 
& 0.055 & 0.265 & 0.104 & 353.824 
& 0.002 & 0.585 & 0.022 & 66.224 
& 0.084 & 0.280 & 0.130 & 368.677 \\

SHAP
& 0.096 & 0.364 & \textbf{0.141} & 563.310 
& \textbf{0.128} & 0.345 & \textbf{0.172} & 651.226 
& 0.002 & 0.857 & 0.024 & 483.520 
& \textbf{0.141} & 0.345 & \textbf{0.204} & 605.251 \\

$L_p$
& 0.072 & 0.345 & 0.116 & 681.217 
& 0.077 & 0.363 & 0.148 & 636.358 
& 0.002 & 0.843 & 0.024 & 428.710 
& 0.077 & 0.331 & 0.120 & 681.624 \\

mRMR
& 0.114 & 0.298 & 0.130 & 575.250 
& 0.126 & 0.313 & 0.165 & 604.981 
& 0.002 & 0.758 & 0.024 & 210.294 
& 0.104 & 0.180 & 0.182 & 394.014 \\

PCA
& 0.076 & 0.327 & 0.122 & 437.053 
& 0.110 & 0.316 & 0.148 & 413.983 
& 0.002 & 0.597 & 0.022 & 69.256 
& 0.085 & 0.298 & 0.131 & 400.975 \\

F-test
& 0.051 & 0.576 & 0.057 & 4120.081 
& 0.046 & 0.551 & 0.059 & 3821.233 
& 0.018 & 0.997 & 0.070 & 709671 
& 0.046 & 0.563 & 0.055 & 3952.806 \\

SHAP, $L_p$
& \textbf{0.122} & 0.379 & \textbf{0.168} & 848.862 
& \textbf{0.133} & 0.409 & \textbf{0.214} & 1080.091 
& 0.002 & 0.909 & 0.024 & 1093.220 
& \textbf{0.168} & 0.392 & \textbf{0.334} & 505.685 \\

SHAP, mRMR
& 0.083 & 0.300 & 0.108 & 570.853 
& 0.093 & 0.336 & 0.127 & 794.904 
& 0.002 & 0.887 & 0.024 & 746.153 
& 0.086 & 0.343 & 0.126 & 426.536 \\

SHAP, PCA
& 0.083 & 0.360 & 0.106 & 582.401 
& 0.101 & 0.378 & 0.172 & 598.437 
& 0.002 & 0.853 & 0.024 & 466.502 
& 0.082 & 0.392 & 0.123 & 451.038 \\

SHAP, F-test
& 0.051 & 0.576 & 0.057 & 4120.081 
& 0.039 & 0.540 & 0.056 & 3793.931 
& 0.018 & 0.997 & 0.070 & 709671
& 0.046 & 0.563 & 0.055 & 3952.806 \\

\bottomrule
\end{tabular}
\end{sidewaystable*}

\end{document}